\documentclass[review]{elsarticle}

\usepackage{hyperref}
\usepackage{epsfig}
\usepackage{graphicx}
\usepackage{amsmath}
\usepackage{amssymb}
\usepackage{caption}
\usepackage{subfigure}
\usepackage{dblfloatfix}
\usepackage{dcolumn}
\usepackage{natbib}
\usepackage{picins}
\journal{}

\begin{document}

\begin{frontmatter}

\title{SA-CNN: Dynamic Scene Classification using Convolutional Neural Networks}

\author[mymainaddress]{Aalok Gangopadhyay}%%\fnref{myfootnote}}
\ead{aalok@iitgn.ac.in}
\author[mymainaddress]{Shivam Mani Tripathi\corref{mycorrespondingauthor}}
\ead{shivam.mani@iitgn.ac.in}
\author[mymainaddress]{Ishan Jindal}
\ead{ijindal@iitgn.ac.in}
\author[mymainaddress]{Shanmuganathan Raman}
\ead{shanmuga@iitgn.ac.in}
\address[mymainaddress]{Indian Institute of Technology Gandhinagar, Gujarat, India}
%%\fntext[myfootnote]{Since 1880.}
\cortext[mycorrespondingauthor]{Corresponding author}
%%\author[mymainaddress,mysecondaryaddress]{Elsevier Inc}
%%\ead[url]{www.elsevier.com}

%%\author[mysecondaryaddress]{Global Customer Service%%

%%\address[mymainaddress]{1600 John F Kennedy Boulevard, Philadelphia}
%%\address[mysecondaryaddress]{360 Park Avenue South, New York}

\begin{abstract}
The task of classifying videos of natural dynamic scenes into appropriate classes has gained lot of attention in recent years. The problem especially becomes challenging when the camera used to capture the video is dynamic. In this paper, we analyse the performance of statistical aggregation (SA) techniques on various pre-trained convolutional neural network(CNN) models to address this problem. The proposed approach works by extracting CNN activation features for a number of frames in a video and then uses an aggregation scheme in order to obtain a robust feature descriptor for the video. We show through results that the proposed approach performs better than the-state-of-the arts for the Maryland and YUPenn dataset.
The final descriptor obtained is powerful enough to distinguish among dynamic scenes and is even capable of addressing the scenario where the camera motion is dominant and the scene dynamics are complex. Further, this paper shows an extensive study on the performance of various aggregation methods and their combinations. We compare the proposed approach with other dynamic scene classification algorithms on two publicly available datasets - Maryland and YUPenn to demonstrate the superior performance of the proposed approach. 

\end{abstract}


\end{frontmatter}

%\linenumbers

%%%%%%%%% BODY TEXT
\section{Introduction}

Consider the video of a natural dynamic scene. The video could have been captured either by a static or a dynamic camera. Given several categories comprising of natural scene videos, we would like to assign the correct category for a given video. This dynamic scene classification problem is more challenging for a moving camera than that for a static camera. 

In the case of images, a lot of significant research has been done to address the problem of scene recognition. Image scene recognition involves classifying an image into one of the several given classes (SUN Database) \cite{xiao2010sun}. Convolution Neural Network (CNN) based approaches have recently dominated the task of image scene classification, obtaining very high accuracy and outperforming other previous state-of-the-art approaches by a significant margin. These approaches have worked remarkably well on several other large scale image datasets with upto thousands of categories. These powerful methods focus on finding appropriate spatial feature descriptors for a given image. Hence, they take into consideration only the spatial description of the scene present in the image.  

\begin{figure}[htbp]
\centering
\captionsetup{justification=centering}
\subfigure[]{\includegraphics[width = 2.0in]{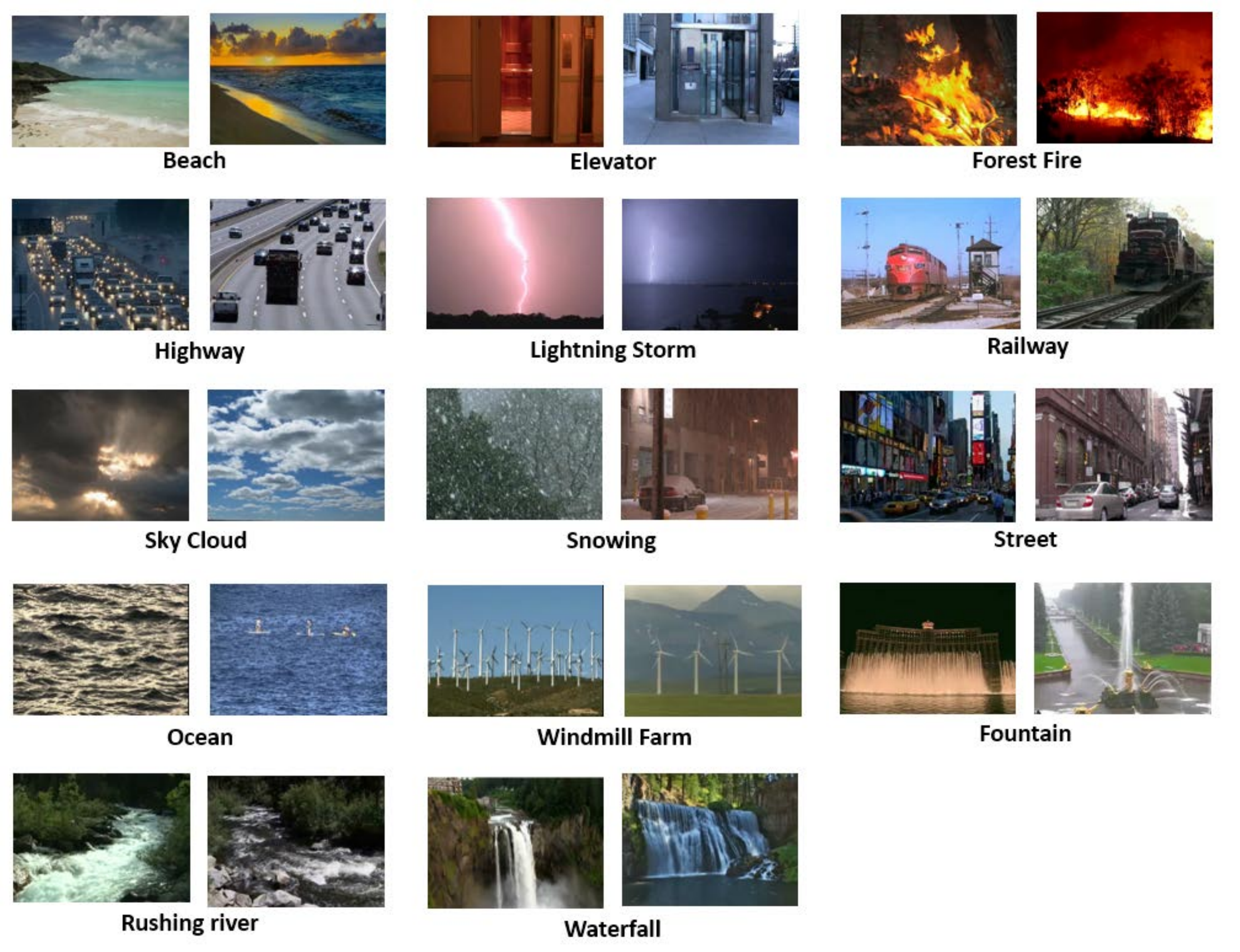}}
\subfigure[]{\includegraphics[width = 2.0in]{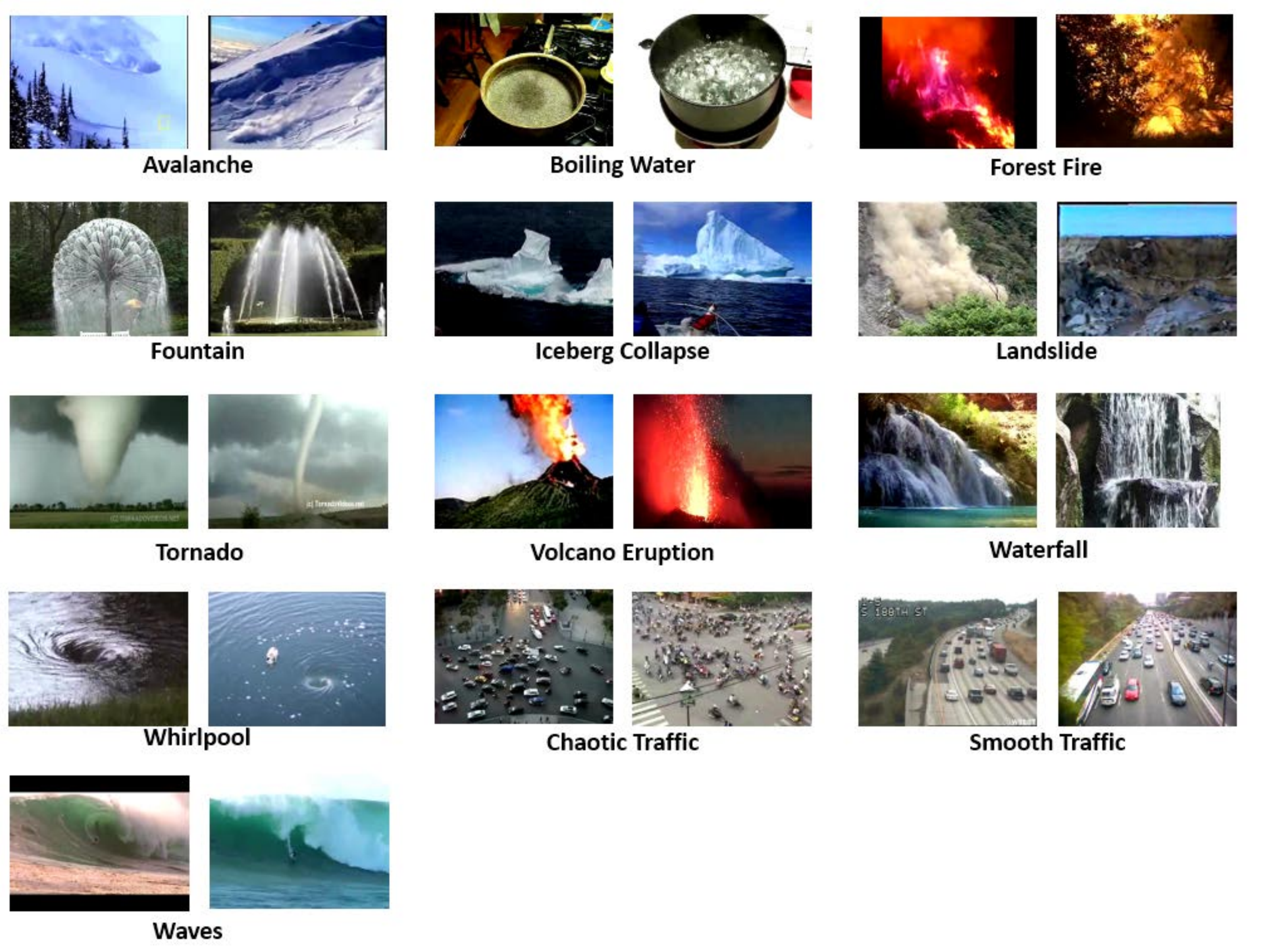}}
\caption{Example frames of classes within (a) YUPenn dataset (left) and (b) Maryland dataset (right).} \label{dataset_preview}
\end{figure}

In contrast to image scene classification where the class labels are based only on the spatial properties, dynamic scene classification tries to classify videos into different categories whose semantic labels is derived from the activities occurring in the scene.  Several examples of dynamic scenes are shown in Figure \ref{dataset_preview}. The dynamic scenes like 'avalanche' is given its class label based on the movement of ice, and not just based on the spatial attributes of the scene. 

The proposed approach is inspired by the unparalleled success of Convolutional Neural Network (CNN) based approaches for various recognition tasks over the past few years. Krizhevesky et al. mentioned the idea of using very large and deep CNN models to classify videos as well \cite{Krizhevsky_imagenetclassification}. But building new architecture for videos and training it on a very large dataset is a complex procedure. However, two recent works on large scale video classification use CNNs to achieve the task by learning features from hundreds of thousands of videos extracted from Youtube or Facebook \cite{karpathy2014large} \cite{tran2014c3d}. In \cite{karpathy2014large} , the proposed architecture was trained on a large collection of sports videos(Sports-1M) for about a month and obtained very good results when tested on UCF-101.   

A number of CNN implementations pretrained on a large database of images are available which are ready to be used for off-the-shelf image feature extraction. We take forward this idea of classification with off-the-shelf  CNN implementation in order to perform dynamic scene classification for videos by using Caffe CNN framework. We utilize several pre-trained models such as AlexNet \cite{jia2014caffe},Places and Hybrid Places model \cite{zhou2014places} for the task. The present approach differs from that of \cite{karpathy2014large} in the sense that the CNN trained on standard image dataset (ImageNet, Places) is used to classify videos of dynamic scenes. This relieves us from training CNN on a new video dataset. We use very simple yet effective tools for dynamic scene classification and show that even common statistical measures can be employed to capture the temporal variation which can be combined with spatial information for dynamic scene videos. We enhance this framework and adapt it for the problem of dynamic scene classification and obtain very high accuracy. 
%These techniques were able achieve scores higher than previous highly sophisticated state of the art techniques using Caffe\cite{jia2014caffe} implementation of CNN. 

It is worth mentioning that all of  the previous dynamic scene classification methods, except Tran et al.'s C3D\cite{tran2014c3d}, relied on using local features and did not exploit very large dataset. The proposed approach explores  three different models of CNN pre-trained on ImageNet (ILSVRC 2012), Places, and Hybrid (combination of both) datasets.  The ImageNet database is largely dedicated to object recognition tasks but not dynamic scene classification tasks \cite{ILSVRCarxiv14}. On the other hand, Places and Hybrid dataset consist fully or partially of scene images, hence we expect them to have more discriminative power for dynamic scene classification task.

The primary contribution of the proposed approach are listed below.
\\ 1. Exploiting pre-trained CNN models and adapting it to the dynamic scene classification task for obtaining high accuracy, \\2. Using common statistical measures to merge spatial features with their temporal variation to arrive at a novel feature descriptor, \\3. We perform comparative study on different off-the-shelf CNN models and compare different pooling strategies, \\4. We design the classification algorithm in such a way that it is highly robust to scene motion as well as camera motion, \\5. We obtain state-of-the-art result on two dynamic scene datasets - Maryland and YUPenn. The increase in classification accuracy is observed to be very high in the case of Maryland dataset.

The rest of the paper is organized as follows. Section 2 describes the relevant works about dynamic scene classification and the CNN literature in detail. Section 3 presents a detailed account of the proposed approach. In section 4, we present results and comparisons of the experiments carried out with complete quantitative analysis. We conclude the paper in section 5 with some suggestions for future work.

\section{Related Work}

A lot of work has been done in the field of scene classification in the past decade. This includes recognizing scenes from single images as well as classifying dynamic scenes from videos. In this section, we shall elaborate on some of the past works which are directly related to the present work.

In the field of single image recognition tasks, bag-of-features based methods were initially prevalent among the research community \cite{Csurka04visualcategorization,Grauman05thepyramid,JDSP10, conf/cvpr/PerronninD07, Wang10locality-constrainedlinear}. These methods were based on the principle of geometric independence or orderless spatial representation. Later, these methods were enhanced by the inclusion of weak geometric information through spatial pyramid matching (SPM)\cite{Schmid06beyondbags}. This method employed histogram intersection at various levels of the pyramids for matching features. However, CNN based approaches have been able to achieve even higher accuracies as observed in some of the recent works \cite{DBLP:journals/corr/GongWGL14,Krizhevsky_imagenetclassification, DBLP:journals/corr/SermanetEZMFL13,DBLP:journals/corr/ZeilerF13}. This sparked a lot of recent research work on architectures and applications of CNNs for visual classification and recognition tasks.

In \cite{Krizhevsky_imagenetclassification}, the CNN architecture consisted of five convolutional layers, followed by two fully connected layers (4096 dimensional) and an output layer. The output from the fifth max-pooling layer was shown to still preserve global spatial information\cite{DBLP:journals/corr/ZeilerF13}. Even the activation features from the fully connected layer were found to be sensitive to global distortions such as rotation, translation, and scaling \cite{DBLP:journals/corr/GongWGL14}. However, they have proven to be very powerful general feature descriptors for high level vision tasks. Several CNN implementations such as DeCAF, Caffe and OverFeat, trained on very large datasets are available for feature extraction to perform image classification tasks \cite{DBLP:journals/corr/DonahueJVHZTD13,jia2014caffe,DBLP:journals/corr/SermanetEZMFL13}. These CNNs, pre-trained on large datasets such as ImageNet, have been efficiently used in scene classification and have achieved high/impressive accuracies \cite{DBLP:journals/corr/GongWGL14} (for example, MOP CNN, OverFeat, etc.).  Also, the ImageNet trained model of these implementations have been shown to generalize well to accommodate other datasets as well \cite{DBLP:journals/corr/DonahueJVHZTD13,DBLP:journals/corr/ZeilerF13}. CNN features obtained from object recognition datasets have also been used for obtaining high accuracy in various high level computer vision tasks \cite{razavian2014cnn}.

On the other hand, research in dynamic scene classification from videos has been dominated by the idea of finding more powerful and robust local spatio-temporal feature descriptors. This is followed by embedding weak global information to find most appropriate representation of the given video. Initially, spatial and temporal feature based approaches such as GIST+HOF and GIST+Chaos were employed to perform dynamic scene classification \cite{MLS09,5539864,vasudevan2013dynamic}. In \cite{MLS09}, it was shown that spatial and temporal descriptors together gave better results than using either of them alone. These methods were built and tested for Maryland (In-the-Wild) dataset introduced by \cite{5539864}.

The spatio-temporal based approaches were introduced by spatio-temporal oriented energies \cite{conf/cvpr/DerpanisLDW12}, which also introduced the YUPenn dataset. The very same work concluded that even relatively simple spatio-temporal feature descriptors were able to achieve consistently higher performance on both YUPenn and Maryland datasets as compared to HOF+GIST and HOF+Chaos approaches. More details for both the dynamic scene datasets have been covered in Section \ref{dataset}. Current state-of-the-art approach, bags of space time energies (BoSE), proposes using a bag of visual words for dynamic scene classification \cite{DBLP:conf/cvpr/FeichtenhoferPW14}. Here, local features extracted via spatio-temporally oriented filters are employed. They are encoded using a learned dictionary and then dynamically pooled. The technique currently holds the highest accuracy on the two mentioned datasets \cite{DBLP:conf/cvpr/FeichtenhoferPW14} amongst all peer-reviewed studies. Recently, a work done by Duran et al. (not peer-reviewed yet)  uses a novel three dimensional CNN architecture  for spatio-temporal classification problems\cite{tran2014c3d}. This technique shows promising results and marginal improvement over current state of art method. Another recent work by Xu et al. used Caffe framework and vectorial pooling (VLAD) to obtain better than state of art performance for the  event detection problem\cite{XuTrecvid2015}. Off the shelf descriptors were used to obtain high score on TRECVID-MED dataset.

\section{Methodology}

\begin{figure}[Ht]
\centering
\captionsetup{justification=centering}
\includegraphics[width=0.9\textwidth]{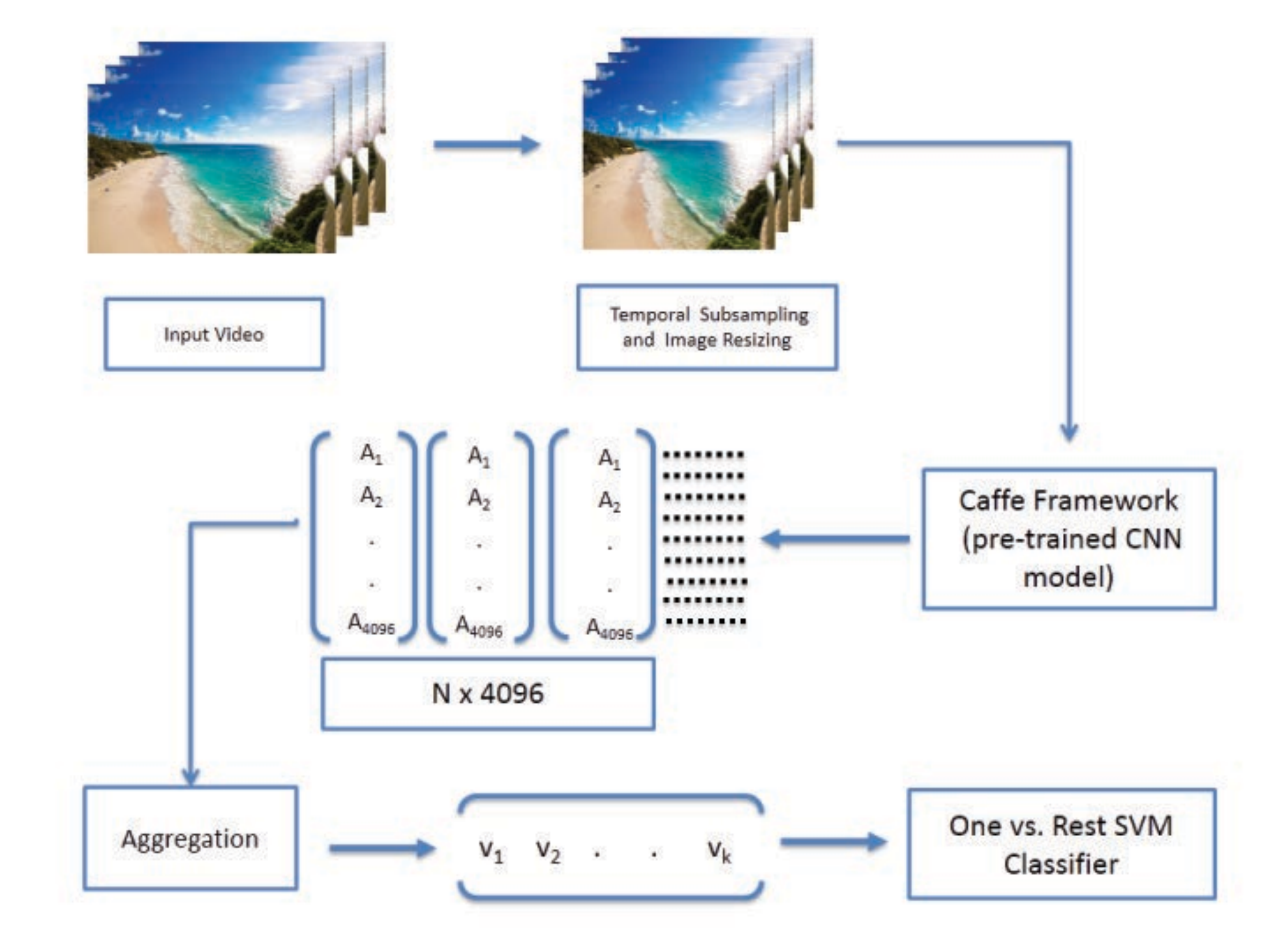}
\caption{The proposed workflow \label{Method}}
\end{figure}

Most of the recent works on dynamic scene classification have focused on dense extraction of spatio-temporal features, followed by feature encoding and pooling strategies to obtain the final feature representation for a video. Several other methods have considered separately extracting spatial and temporal features and then combining them to obtain the final feature representation for a given video. However, we use a different approach here. Given a video, we first extract spatial feature descriptors for a chosen number of frames. After that, we use aggregation strategies to obtain information about the temporal variation of the spatial features. Using this information, we arrive at the final feature descriptor for a given video to be classified. The entire process has been outlined in Figure \ref{Method}. In this section, we shall describe the proposed approach in more detail.

\subsection{Pre-trained CNN Models}
For the feature extraction purpose, we experiment with the following three pre-trained models based on the Caffe framework:

(1) \textbf{AlexNet} \cite{jia2014caffe}: This model has been trained on ILSVRC2012 (ImageNet) with 1000 categories and over a million images\cite{ILSVRCarxiv14}. Please note that this model has minor variations from the model mentioned in \cite{Krizhevsky_imagenetclassification} 

(2) \textbf{Places205-CNN}\cite{zhou2014places} : This model has been trained on 205 scene categories of Places Database with approximately 2.5 million images.

(3) \textbf{Hybrid-CNN}\cite{zhou2014places} : This model has been trained on 1183 categories (205 scene categories from Places Database and 978 object categories from the train data of ILSVRC2012) with 3.6 million images.

\subsection{Feature Extraction from Frames}
To start with, we extract feature descriptors for the frames of a video using spatial information only. Given a video $V_{k}$ containing a total of $N_{0k}$ frames, we select $N_{k}$ frames that are linearly spaced in the interval $[1,N_{0k}]$. An important thing to note here is that the temporal distance between consecutive frames in the set of $N_{k}$ frames differ from video to video, as there is a lot of variation in the frame rate and the total video duration in the datasets. After selection, each of these $N_{k}$ frames are resized appropriately (AlexNet model uses a resolution of $256\times 256$ while the other two models use $227\times 227$) . Then we extract the CNN activation features for each of these frames using the above mentioned pretrained CNN Caffe implementations.
%Also, we do not perform any additional fine tuning to the CNN model in order to %keep the model generic. 
For each of the frame $f_{i}$ taken as input, we take the output of either the sixth(fc6) or the seventh(fc7) fully connected layer of the CNN (post ReLU transformation) and obtain the 4096 dimensional feature descriptor $X_{i}$. Thus, after the feature extraction step, we obtain an  $N_{k} \times 4096$ dimensional matrix $X$ for the video $V_{k}$. This matrix contains the information about how each of these 4096 features evolve with time and hence can be exploited to extract the temporal properties. We have 4096 time curves each of length Nk. Ideally we would like to use a feature descriptor for each of these curves that captures the temporal variations in a robust way. But such an ordered temporal descriptor would make the final descriptor for the video very huge, since there are 4096 such curves. Hence, rather than using the ordered properties, we use temporally orderless statistics for each curve. We show that simply using the first few moments yield very high accuracy for both the datasets.

\subsection{Aggregation}
From the previous step, we obtain a set of 4096 dimensional vectors, $\{X_1, ... , X_K\}$, each of which contains rich spatial information and represents a single frame in the given video. In this step, we combine these $K$ vectors $\{X_1, ... , X_K\}$ in time in order to capture the temporal statistics of the spatial features . We do this to induce a certain degree of temporal invariance and extract temporally order-less properties. For this, we consider two strategies to aggregate these spatial descriptors temporally.

\textbf{Statistical Aggregation (SA):} The simplest method is to use statistical measures like moments, for $M=N_{k}$ instances for each of the 4096 dimensions. Let $X_{ij}$ denote the $i^{th}$ instance of $j^{th}$ dimension of the feature descriptor, where $i \in \{1,2..,M\}$ and $j \in \{1,2 ..,4096\}$. Then we use the following statistical measures to aggregate $M$ instances to get the final feature descriptors.
\begin{enumerate}
  \item Mean: $\hat{\mu} = (\mu_1, \mu_2, ..., \mu_{4096})$ where: $ {\mu}_{j} = \frac{1}{M}\sum\limits_{i=1}^M  X_{ij}  $
  \item Standard Deviation: $\hat{\sigma} = (\sigma_1, \sigma_2, ..., \sigma_{4096})$ where: ${\sigma}_{j} = \sqrt{\frac{1}{M} \sum_{i=1}^M (X_{ij} - {{\mu}_{j}})^2}$
  \item Skewness: $\hat{\gamma} = (\gamma_1, \gamma_2, ..., \gamma_{4096})$ where:$ {\gamma}_{j} = \frac{\operatorname{E} \left [(X_{j} - \mu_{j})^3 \right ]}{\sigma_j^3}$
  \item Kurtosis: $\hat{\kappa} = (\kappa_1, \kappa_2, ..., \kappa_{4096})$ where: $ {\kappa}_{j} =  \frac{\operatorname{E} \left [(X_j - \mu_j)^4 \right ]}{\sigma_j^4} $
  \item Max: $\hat{m} = (m_1, m_2, ..., m_{4096})$ where: $ {m}_{j} = max \{X_ij$ where $i \in (1,2..,M)\} $
\end{enumerate} 

Please note that each of $\hat{\mu}, \hat{\sigma}, \hat{\gamma}$ and $\hat{\kappa}$ are 4096 dimensional vectors. For classification, we can consider each one of these moments individually or their various combinations (concatenation) to get the final feature descriptor of the video. Details of the combinations used have been covered in section \ref{4.0}. Apart from using the moments, we also evaluate the performance of the classifier on using max-pooling for temporal aggregation.

Vector of Locally Aggregated Descriptors (VLAD) is one of the most popular vectorial pooling strategy in many different computer vision tasks. Developed after bag-of-words (boW) feature encoding, VLAD was designed as an compact enhanced feature descriptor \cite{jegouVlad2010}. VLAD was shown to outperform BoW in many vision related tasks. It primarily differs from BoW model in the sense that instead of recording number of vectors assigned to each cluster center, it records difference from different centres and accumulates them. We used VLFeat implementation of VLAD for our experiments \cite{vlfeat}. The pipeline to obtain final VLAD encoded features for our method is as follows.

First, we collect all 4096-D vectors from all videos, reduce them to $D^{'}$ dimensions using PCA. We store this transformation. Then, we learn a codebook of  $k$ cluster centres by running K-Means++ on the dimension reduced  set obtained in first step. An encoding step follows this. Given M 4096-D vectors ($\{X_1,X_2,..,X_M\}$) for a video, we first reduce it to M $D^{'}$-dim vectors using stored PCA transformation (with whitening). Let $\{ {X^{'}}_1, {X^{'}}_2 ,.., {X^{'}}_M \}$ be reduced vectors for a given video. Let $\{ c_1, c_2 ,.., c_k \}$ be the obtained cluster centres. Let $q_{ik}$ be the strength of the association of vector $X_i$ to cluster $c_k$ obtained through KDTree quantization with hard assignment. Then VLAD encodes feature x by considering the residuals:

\begin{center}
$v_k = \sum\limits_{i=1}^M q_{ik}( X_i-c_k)$
\end{center}
We do this for each of the $k$ cluster center, and all the resulting residuals ($v_k$'s) are accumulated and concatenated to obtain single $kD^{'}$ dimensional feature vector for the video.

\subsection{Classification}
The video descriptor obtained from aggregation step are fed to Linear SVM Classifier. More details classification methodology is covered in section\ref{4.0}.

A good method to measure the merits of aggregation scheme listed in section 3.3 would be to compare it against results obtained with a majority voting classification scheme. We implement this scheme as follows:
Let $X = \{X_1,X_2,..,X_M\}$ be the set of 4096-D feature vectors obtained from CNN for M linearly space frames of a video (from section 3.2). Then a majority voting classifier $C$ classifies a video (represented by set X) into one of given classes via function:
\begin{center}
$ C(X) = mode\{ h(X_1), h(X_2) ,.., h(X_M)\} $
\end{center}
where $h$ evaluates single frame via one-vs-rest classification rules.

In the next section, we shall explain the various experiments we carried out and show their results. We also show the comparison with other competing methods to emphasize the significance of the contribution in this work.

\section{Results and Discussion}\label{4.0}

\subsection{Dataset}\label{dataset}
We evaluate our method on the following two datasets. 

1. \textbf{YUPenn} (Stabilized Dataset) : This dataset contains 14 classes with 30 videos in each class making it a total of 420 videos. Each video contains around 145 frames on an average, with the frame rate not being the same for each video \cite{conf/cvpr/DerpanisLDW12}. 

2. \textbf{Maryland} (In-the-Wild Dataset) : This dataset contains 13 classes with 10 videos in each class making it a total of 130 videos. Each video has the same frame rate of 30 fps. But the duration of the videos and hence the total number of frames varies a lot in the dataset \cite{MLS09}.

Both the datasets have large variation in illumination, image scale, camera viewpoint, frame resolution, duration of the video, etc. The videos in Maryland dataset contain large camera motion and scene cuts, whereas those in YUPenn involve static camera and hence contain no camera motion. All the above factors result in large intraclass variations which make dynamic scene classification a challenging task.

\subsection{Results}

As mentioned in the proposed approach, we first choose one of the three pre-trained CNN model for feature extraction. Then given a video $V_{k}$,  we extract the 4096 dimensional activation features from either the fc6 or the fc7 layer for each of the $N_{k}$ frames chosen from the video. Thus we obtain $N_{k}$ such 4096 dimensional feature vectors for the video $V_{k}$. To find the final vector representation for each video, we perform temporal aggregation using statistical moments, max pooling or vectorial pooling. After obtaining the feature vector representation for each video in the dataset we perform multi-class classification. For classification, we use ‘one-vs-rest’ SVM with the leave-one-video-out (LOVO) method as done in previous works \cite{DBLP:conf/cvpr/FeichtenhoferPW14}. For the implementation of SVM, we use the LIBSVM library\cite{CC01a}. It is found that in the case of Maryland dataset, linear kernel gives best results and in the case of YUPenn, histogram intersection kernel(HIK) works the best.

First, we conduct an experiment to find out which of the three pre-trained model is best suited for the dynamic scene classification task. For this, we evaluate the classification accuracies obtained using (a) mean pooling, (b) max pooling and (c) combination of the first four moments. We evaluate for both the fc6 and the fc7 layer. In this experiment all the frames of a video are used for aggregation. After aggregation, the final features for each video are normalised and then classification is performed. The results obtained are mentioned in Table \ref{Maryland_accuracy} (for Maryland) and Table \ref{YUPenn_accuracy} (for YUPenn). From the tables it is clearly visible that HYBRID-CNN outperforms the other two models in all the three cases. This is owing to the fact that hybrid model has been pre-trained on both objects and scenes, and hence is more suitable for dynamic scene classification.
It can also be observed that the accuracies obtained using fc7 are greater than or equal to the ones obtained using fc6. Hence, in all the further experiments we use the fc7 layer of the HYBRID-CNN model for the feature extraction task in both Maryland and YUPenn.
\begin{table} [ht]
{\small
%	\begin{center}
\resizebox{0.80\textwidth}{!}{\begin{minipage}{0.9\textwidth}

		\begin{tabular}{  m{3.3cm} |  c  c | c  c | c  c  }

\hline 

\rule{0pt}{4ex}    &\multicolumn{2}{c|}{ALEXNET} &\multicolumn{2}{c|}{PLACES} &\multicolumn{2}{|c}{HYBRID} \\ \hline \rule{0pt}{4ex}  
Aggregation & fc6 & fc7 & fc6 & fc7 & fc6 & fc7\\ \hline \rule{0pt}{4ex}  
Mean    &89.23    &90.00    &89.23    &86.92    &\textbf{90.77}    &\textbf{90.77} \\ \hline\rule{0pt}{4ex}  
Max    &88.46    &87.69    &87.69    &89.23    &91.54    &\textbf{93.08} \\
\hline\rule{0pt}{4ex} 
Mean + S.D. + Skew + Kurtosis    &88.46    &91.54    &87.69    &85.38    &\textbf{93.85}    &\textbf{93.85}\\ \hline
\end{tabular}
	\end{minipage}}
	\caption{Comparison of different models for MARYLAND }
	\label{Maryland_accuracy}
%	\end{center}
	}
\end{table}

\begin{table} [ht]
{\small
%	\begin{center}
\resizebox{0.80\textwidth}{!}{\begin{minipage}{0.9\textwidth}

		\begin{tabular}{  m{3.3cm} |  c  c | c  c | c  c  }

\hline 

\rule{0pt}{4ex}    &\multicolumn{2}{c|}{ALEXNET} &\multicolumn{2}{c|}{PLACES} &\multicolumn{2}{|c}{HYBRID} \\ \hline \rule{0pt}{4ex}  
Aggregation & fc6 & fc7 & fc6 & fc7 & fc6 & fc7\\ \hline \rule{0pt}{4ex}  
Mean    &96.90    &96.90    &96.43    &97.38    &\textbf{98.10}    &\textbf{98.10} \\ \hline\rule{0pt}{4ex}  
Max    &96.90    &96.43    &96.67    &96.67    &97.62    &\textbf{97.86} \\
\hline\rule{0pt}{4ex} 
Mean + S.D. + Skew + Kurtosis    &96.67    &97.14    &96.19    &96.43    &97.62    &\textbf{98.33}\\ \hline
\end{tabular}
	\end{minipage}}
	\caption{Comparison of different models for YUPenn }
	\label{YUPenn_accuracy}
%	\end{center}
	}
\end{table}

To analyse the performance of the statistical moments, we initially utilize all the frames, that is, we set : $N_{k} = N_{0k}$. We later show that we get good results even for smaller values of $N_{k}$ . Table \ref{statistical accuracies} depicts the results obtained on using statistical moments and their various possible combinations. 

\begin{table}[htbp]
{
	\begin{center}
		\begin{tabular}{ m{3.3cm}| c | c | c }
\hline

%\multicolumn{3}{|c||}{Stastical Agg.}\\ \hline
 
Statistical Measures 	& Dim		& Yupenn      & Maryland   \\ \hline

Mean					&4,096		 & 98.10 	  & 90.77  		\\ 
S.D.					&4,096		 & \textbf{98.57}  	  & 92.31   	\\ 
Skew					&4,096		 & 95.00  	  & 76.92   	\\ 
Kurtosis				&4,096		 & 95.00  	  & 58.46   \\ \hline
Mean$+$S.D.		     	&8,192		 & \textbf{98.57}  	  & 93.08   \\ 
Mean$+$Skew				&8,192		 & 98.10  	  & 93.08   \\ 
Mean$+$Kurtosis			&8,192		 & 98.10  	  & 93.08   \\ 
S.D.$+$Skew				&8,192		 & 97.86  	  & 92.31   \\ 
S.D.$+$Kurtosis			&8,192		 & 97.62  	  & 92.31   \\ 
Skew$+$Kurtosis			&8,192		 & 96.19 	  & 70.00   \\ \hline
Mean$+$S.D.$+$Skew		&12,288		 & 98.10  	  & \textbf{93.85}  \\ 
Mean$+$S.D.$+$Kurtosis	&12,288		 & 98.10  	  & \textbf{93.85}   \\ 
Mean$+$Skew$+$Kurtosis	&12,288 		 & 97.86  	  & \textbf{93.85}   \\ 
S.D.$+$Skew$+$Kurtosis	&12,288		 & 97.86  	  & 93.08   \\ \hline
Mean$+$S.D.$+$Skew $+$Kurtosis  &16,384	 & \textbf{98.33}  	  & \textbf{93.85}   \\ \hline

		\end{tabular}
\caption{Accuracy obtained for the various moments and their combinations (HYBRID-CNN fc7): The first block (a) shows the results of using the first four moments individually Each of these vectors have a dimension of 4096. In the second block (b), doublet combinations of the moments are obtained by concatenation. Each of these vectors have a dimension of 8192. The third block (c), triplet combinations are considered, each of them having a dimension of 12288. The fourth block (d) shows the result obtained on combining all the four moments resulting in a vector of dimension 16384.}
\label{statistical accuracies}
	\end{center}
	}
\end{table}

It is surprising to see that apart from mean, even the other statistical moments, such as S.D., skew and kurtosis perform well when considered individually (as seen in block (a) of table \ref{statistical accuracies}).This means that the temporal statistics of the 4096 dimensional vector is highly similar for videos of the same class. This indicates that various types of probability based approaches can be explored for obtaining a very powerful descriptor for the videos. Quick observation of the table reveals that concatenation of all the four moments: mean, standard deviation (S.D.), skew and kurtosis consistently gives high scores on both the datasets. Also, all the feature descriptors which contain either of mean, S.D or their concatenation  perform well.

%\rowcolors{1}{}{lightgray}

We obtained average classification accuracy of 80\% in YUPenn dataset and 16\% in Maryland dataset while using VLAD. This poor performance is due to the fact that the Maryland dataset contains both camera and scene motion and also the number of frames per video also varies drastically. However, we are able to get an accuracy of 85\% in Maryland dataset through just dimensionality reduction of 4096-D feature using PCA followed by simple average pooling and classification using SVM.

Since the time taken for computing the final feature vector for a given video $V_{k}$ largely depends on the value of $N_{k}$ , it is very important to understand how the accuracy of the classifier varies with $N_{k}$. Moreover, to make the computation time independent of the duration or the frame rate of the video, it would be better to choose the same value of $N_{k}$ for all videos. Hence, we take same $N_{k} = N$ for all videos. For obtaining the relation between the accuracy and the number of frames selected from each video, we perform a multiple trials based experiment. In this simple experiment, we evaluate our method for the following values of $N \in \{1,2,3,5,10,15,20,30,40,50,60\}$. For each of these values of $N$, one-vs-rest SVM with leave-one-video-out (LOVO) is performed 18 times. In each trial $N$ frames are randomly chosen from the video and the 4096 dimensional feature vectors are aggregated using their temporal mean. We evaluate this behavior only for mean, since, for evaluating the accuracy in case of higher moments we need substantial number of frames and hence can't test them for very low number of frames like 1,2,3 and 5.

\begin{figure*}[ht]
\centering
\captionsetup{justification=centering}
\subfigure[]{\includegraphics[width=0.42\textwidth]{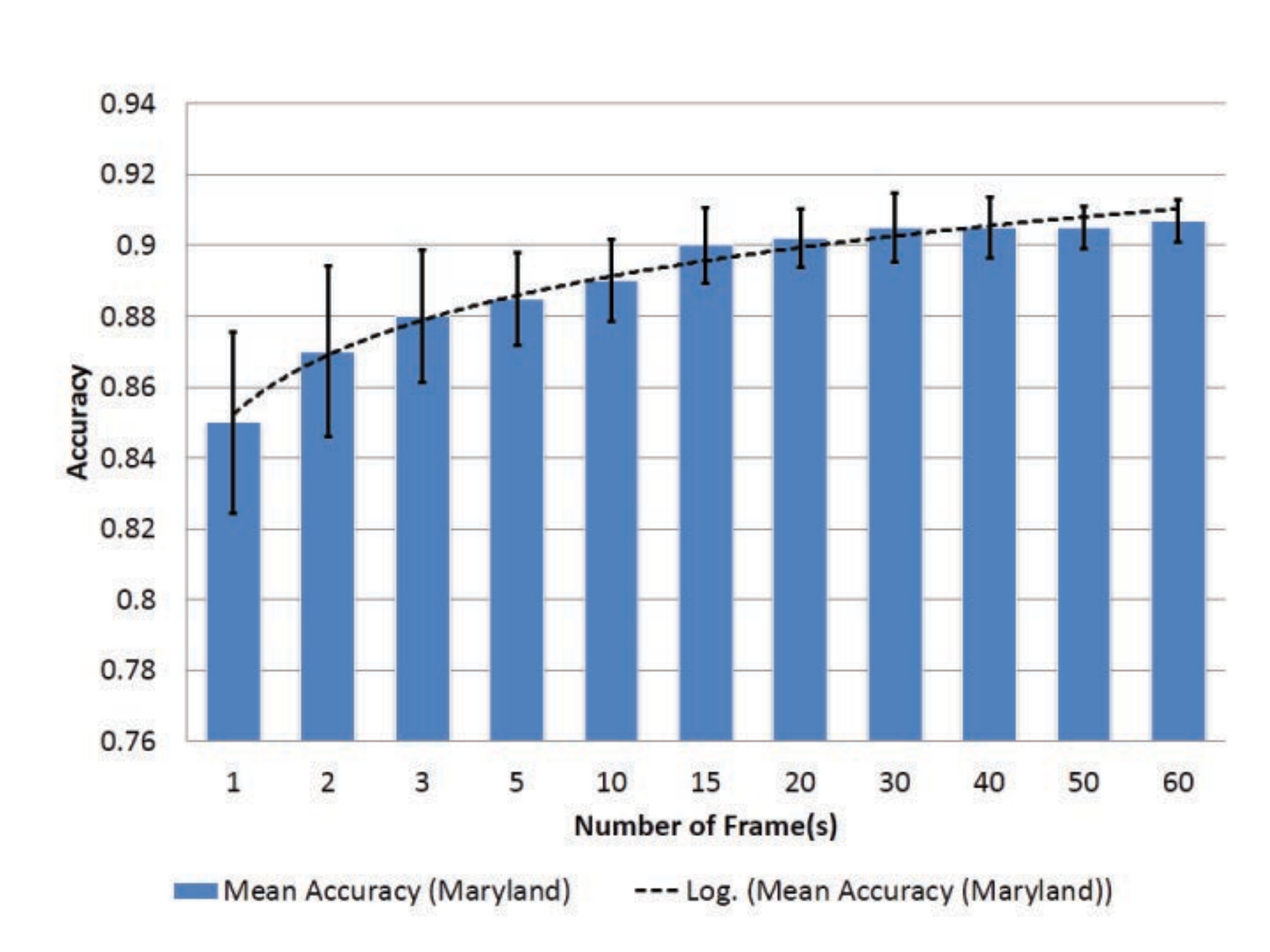}}
\subfigure[]{\includegraphics[width=0.42\textwidth]{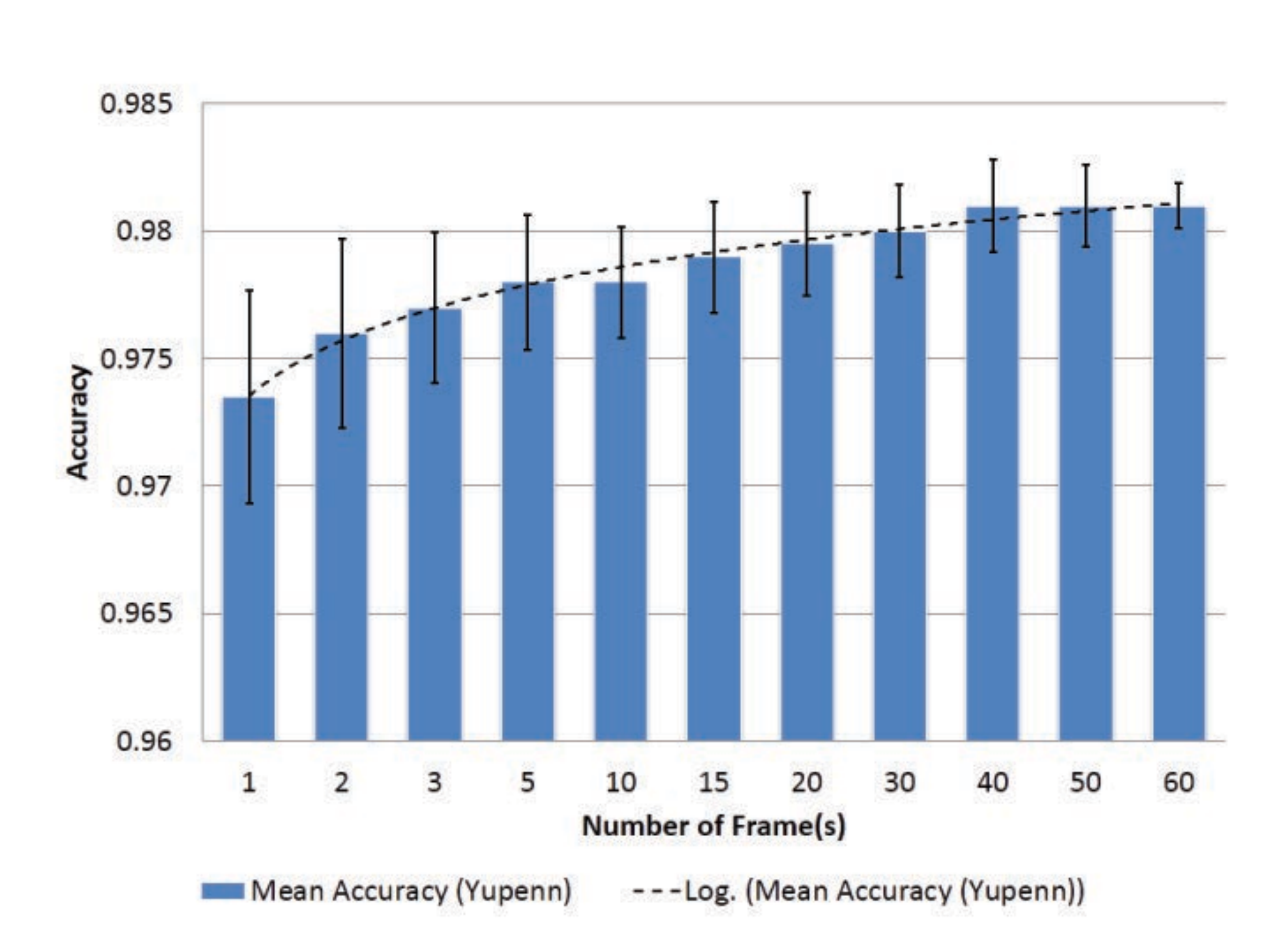}}
\caption{Accuracy vs number of frames using mean aggregation. Here x-axis denotes the number of frames used for aggegation and the y-axis denotes the average accuracy obtained for 18 trials in each case.(a) Maryland (left) and (b) YUPenn (right)} \label{frame_vs_accuracy}
\end{figure*}

%\rowcolors{1}{}{lightgray}
\begin{table} 
{
\resizebox{0.56\textwidth}{!}{
\begin{minipage}{0.9\textwidth}
\begin{flushleft}   
	%\begin{center} 
		\begin{tabular}{  c  p{1cm}  p{1.2cm}  p{1cm}  c  c  c  c p{1.3cm} } 
\hline \rule{0pt}{4ex}  

Class  & HOF + GIST \cite{MLS09} & Chaos + GIST \cite{5539864}& SOE \cite{conf/cvpr/DerpanisLDW12} & SFA \cite{theriault2013dynamic} & CSO \cite{feichtenhofer2013spacetime} & BoSE \cite{DBLP:conf/cvpr/FeichtenhoferPW14}  &C3D\cite{tran2014c3d}  & SA-CNN	[$\hat{\mu}, \hat{\sigma}, \hat{\gamma}, \hat{\kappa}$]\\ \hline\rule{0pt}{4ex}  

Avalanche       & 20 	&60 &40 &60 &60 &60  &NA  & 100\\ 
Boiling Water	& 50  	&60 &50 &70 &80 &70	 &NA & 90\\ 
Chaotic Traffic	 & 30  	&70 &60 &80 &90 &90	  &NA & 100\\ 
Forest Fire		 & 50  	 &60 &10 &10 &80 &90   &NA & 100\\ \
Fountain		 & 20  	 &60 &50 &50 &80 &70    &NA & 90\\ 
Iceberg Collapse & 20  	 &50 &40 &60 &60 &60    &NA &100 \\ 
Landslide		 & 20  	 &30 &20 &60 &30 &60   &NA  & 90\\ 
Smooth Traffic	 & 30  	  &50 &30 &50 &50 &70   &NA  & 90\\ 
Tornado			 & 40  	 &80 &70 &70 &80 &90  &NA & 90\\ 
Volcanic Eruption & 20 	 &70 &10 &80 &70 &80  &NA  & 90\\ 
Waterfall		 & 20  	 &40 &60 &50 &50 &100  &NA   & 100\\ 
Waves			 & 80  	  &80 &50 &60 &80 &90   &NA  & 90\\ 
Whirlpool	 	 & 30  	 &50 &70 &80 &70 &80  &NA & 90\\ \hline\rule{0pt}{4ex}  
Overall			 &33.08 &58.46 &43.08 &60.00 &67.69 &77.69  &77.69  & \textbf{93.85} \\ \hline

\end{tabular}

	%\end{center}
    \end{flushleft}   
	\end{minipage} }
	\caption{Comparison of classification scores for various methods of Dynamic Scene Recognition on Maryland dataset.  }
	\label{maryland_accuracy}
	}
\end{table}

\begin{table} 
{
\resizebox{0.60\textwidth}{!}{\begin{minipage}{0.9\textwidth}
\begin{flushleft}   
	%\begin{center}
		\begin{tabular}{  c  p{1cm}  p{1.2cm}  p{1cm}  c  c  c  c p{1.3cm} }
\hline \rule{0pt}{4ex}  

Class  &  HOF + GIST \cite{MLS09} & Chaos + GIST \cite{5539864}& SOE \cite{conf/cvpr/DerpanisLDW12} & SFA \cite{theriault2013dynamic} & CSO \cite{feichtenhofer2013spacetime} & BoSE \cite{DBLP:conf/cvpr/FeichtenhoferPW14} &C3D\cite{tran2014c3d} & SA-CNN	\\ \hline\rule{0pt}{4ex}  

Beach  & 87  & 30  & 93  & 93  & 100  & 100  &NA & 97\\   
Elevator  & 87  & 47  & 100  & 97  & 100  & 97  &NA &100\\  
Forest Fire  & 63  & 17  & 67  & 70  & 83  & 93  &NA  &100\\  
Fountain  & 43  & 3  & 43  & 57  & 47  & 87   &NA   &100\\  
Highway  & 47  & 23  & 70  & 93  & 73  & 100  &NA  &100\\
Lightning Storm  & 63  & 37  & 77  & 87  & 93  & 97 &NA & 93\\
Ocean  & 97  & 43  & 100  & 100  & 90  & 100 &NA   & 100\\
Railway  & 83  & 7  & 80  & 93  & 93  & 100 &NA & 100\\
Rushing River  & 77  & 10  & 93  & 87  & 97  & 97 &NA & 100\\
Sky-Clouds  & 87  & 47  & 83  & 93  & 100  & 97  &NA &100\\
Snowing  & 47  & 10  & 87  & 70  & 57  & 97  &NA  &97\\
Street  & 77  & 17  & 90  & 97  & 97  & 100 &NA &100\\
Waterfall  & 47  & 10  & 63  & 73  & 77  & 83  &NA   &97\\
Windmill Farm  & 53  & 17  & 83  & 87  & 93  & 100 &NA &93\\ \hline\rule{0pt}{4ex}  
Overall  & 68.33  & 22.86  & 80.71  & 85.48  & 85.95  & 96.19 &96.67 &\textbf{98.33}\\ \hline
\end{tabular}

	%\end{center}
    \end{flushleft}   
	\end{minipage}}
	\caption{Comparison of classification scores for various methods of Dynamic Scene Recognition on YUPenn dataset. }
	\label{yupenn_accuracy}
	}
\end{table}

\begin{table} 
{
\begin{center}
\resizebox{0.7\textwidth}{!}{\begin{minipage}{1.0\textwidth}
\begin{flushleft}   
	
		\begin{tabular}{ c | c@{\hskip 0.05in} c@{\hskip 0.05in} c@{\hskip 0.05in} c@{\hskip 0.05in} c@{\hskip 0.05in} c@{\hskip 0.05in} c@{\hskip 0.05in} c@{\hskip 0.05in} c@{\hskip 0.05in} c@{\hskip 0.05in} c@{\hskip 0.05in} c@{\hskip 0.05in} c@{\hskip 0.05in} c@{\hskip 0.05in} c  }
% \rule{0pt}{4ex}  
    &\rotatebox{90}{beach}    &\rotatebox{90}{fountain}    &\rotatebox{90}{ocean}    &\rotatebox{90}{rushing river}    &\rotatebox{90}{street}    &\rotatebox{90}{elevator}    &\rotatebox{90}{highway}    &\rotatebox{90}{railway}    &\rotatebox{90}{sky clouds}    &\rotatebox{90}{waterfall}    &\rotatebox{90}{forest fire}    &\rotatebox{90}{lightning storm}    &\rotatebox{90}{snowing}    &\rotatebox{90}{windmill farm} \\ \hline
beach    & 29    &0    &0    &1    &0    &0    &0    &0    &0    &0    &0    &0    &0    &0 \\
fountain    &0    &30    &0    &0    &0    &0    &0    &0    &0    &0    &0    &0    &0    &0 \\
ocean    &0    &0    &30    &0    &0    &0    &0    &0    &0    &0    &0    &0    &0    &0 \\
rushing river    &0    &0    &0    &30    &0    &0    &0    &0    &0    &0    &0    &0    &0    &0 \\
street    &0    &0    &0    &0    &30    &0    &0    &0    &0    &0    &0    &0    &0    &0 \\
elevator    &0    &0    &0    &0    &0    &30    &0    &0    &0    &0    &0    &0    &0    &0 \\
highway    &0    &0    &0    &0    &0    &0    &30    &0    &0    &0    &0    &0    &0    &0 \\
railway    &0    &0    &0    &0    &0    &0    &0    &30    &0    &0    &0    &0    &0    &0 \\
sky clouds    &0    &0    &0    &0    &0    &0    &0    &0    &30    &0    &0    &0    &0    &0 \\
waterfall    &0    &0    &0    &1    &0    &0    &0    &0    &0    &29    &0    &0    &0    &0 \\
forest fire    &0    &0    &0    &0    &0    &0    &0    &0    &0    &0    &30    &0    &0    &0 \\
lightning storm    &0    &0    &0    &1    &0    &0    &0    &0    &0    &0    &0    &28    &1    &0 \\
snowing    &0    &0    &0    &1    &0    &0    &0    &0    &0    &0    &0    &0    &29    &0 \\
windmill farm    &0    &0    &0    &0    &0    &0    &0    &0    &0    &0    &0    &0    &2    &28 \\ \hline

\end{tabular}

    \end{flushleft}   
	\end{minipage}}
	\caption{Confusion matrix for YUPenn dataset. }
	\label{yupenn_confusion}
\end{center}	
	}
\end{table}

\begin{table} 
{
	\begin{center}
\resizebox{0.7\textwidth}{!}{\begin{minipage}{1.0\textwidth}

\begin{flushleft}

		\begin{tabular}{ c | c@{\hskip 0.05in} c@{\hskip 0.05in} c@{\hskip 0.05in} c@{\hskip 0.05in} c@{\hskip 0.05in} c@{\hskip 0.05in} c@{\hskip 0.05in} c@{\hskip 0.05in} c@{\hskip 0.05in} c@{\hskip 0.05in} c@{\hskip 0.05in} c@{\hskip 0.05in} c@{\hskip 0.05in} c@{\hskip 0.05in}}

 \rule{0pt}{4ex}  

  &\rotatebox{90}{avalanche}  &\rotatebox{90}{forest fire}  &\rotatebox{90}{landslide}  &\rotatebox{90}{tornado}  &\rotatebox{90}{waves}  &\rotatebox{90}{boiling wate}  &\rotatebox{90}{fountain}  &\rotatebox{90}{volcano erruption } &\rotatebox{90}{whirlpool}  &\rotatebox{90}{chaotic traffic}  &\rotatebox{90}{iceberg collapse}  &\rotatebox{90}{smooth traffic } &\rotatebox{90}{waterfall} \\ \hline

avalanche  &10 &0  &0  &0  &0  &0  &0  &0  &0  &0  &0  &0  &0 \\

forest fire  &0  &10  &0  &0  &0  &0  &0  &0  &0  &0  &0  &0  &0 \\

landslide  &0  &0  &9  &0  &1  &0  &0  &0  &0  &0  &1  &0  &0 \\

tornado  &0  &1  &0  &9  &0  &0  &0  &0  &0  &0  &0  &0  &0 \\

waves  &1  &0  &0  &0  &9  &0  &0  &0  &0  &0  &0  &0  &0 \\

boiling water  &0  &0  &0  &0  &0  &9  &0  &0  &1  &0  &0  &0  &0 \\

fountain  &0  &0  &0  &0  &0  &0  &9  &0  &0  &0  &1  &0  &0 \\

volcano erruption  &0  &1  &0  &0  &0  &0  &0  &9  &0  &0  &0  &0  &0 \\

whirlpool  &0  &0  &1  &0  &0  &0  &0  &0  &9  &0  &0  &0  &0 \\

chaotic traffic  &0  &0  &0  &0  &0  &0  &0  &0  &0  &10  &0  &0  &0 \\

iceberg collapse  &0  &0  &0  &0  &0  &0  &0  &0  &0  &0  &10  &0  &0 \\

smooth traffic  &0  &0  &0  &0  &0  &0  &0  &0  &0  &1  &0  &9  &0 \\

waterfall  &0  &0  &0  &0  &0  &0  &0  &0  &0  &0  &0  &0  &10 \\ \hline

\end{tabular}
    \end{flushleft}
	\end{minipage}}

	\label{maryland_confusion}
	\end{center}	
}
\caption{Confusion matrix for Maryland dataset. }
\end{table}

The graph in figure \ref{frame_vs_accuracy} depicts how the average accuracy obtained for mean aggregation varies with the number of frames. As expected the accuracy increases as the number of frames increase and saturates around 90.7\% for Maryland and around 98.1\% for YUPenn. As the value of N increases, the variation in accuracy across the trials decreases. It is apparent from the graph that, for smaller values of N, the accuracy is very sensitive to which frame is randomly chosen. Thus such a high fluctuation in accuracy is observed. Also, it shows that the confidence in accuracy increases as we take more frames and then saturates roughly after $N > 30$ on both the datasets. From figure \ref{frame_vs_accuracy} we conclude that choosing $N = 30$ is sufficient for getting high accuracy with high confidence. Choosing this optimal value of the number of frames balances out computation time and accuracy.

An interesting thing to note is that, in the case of YUPenn, even for $N=1$, the accuracy is very high (~97.35\%), which is close to the accuracy for larger values of $N$. However, in the case of Maryland, the accuracy improves significantly as $N$ increases. Considering the fact that in Maryland there is lot of camera motion, performing mean aggregation significantly improves the performance as compared to a single frame. But in YUPenn as the camera motion is negligible, even a single frame is very powerful and taking mean results in a slight improvement. This indicates that the simple aggregation scheme robustly handles the effect of camera motion in the videos. We also performed classification based on majority voting scheme explained earlier. We obtained the average accuracy of 90.77\% for Maryland dataset and 97.14\% for YUPenn dataset for 10 linearly spaced frames per video. 
 
We compare the results obtained using statistical aggregation scheme ($\hat{\mu} + \hat{\sigma} + \hat{\gamma} + \hat{\kappa}$) to previous methods for Maryland(Table \ref{maryland_accuracy}) and YUPenn(Table \ref{yupenn_accuracy}). The proposed approach shows the outstanding performance over current state-of-the-art methods (BoSE \& C3D) for the Maryland dataset, with a leap of 16.16\%. The classes iceberg collapse and avalanche witness largest improvement over previous best performing techniques. On other classes, the statistical aggregation is either at par or ahead in terms of classification accuracy. The state of the art results for the YUPenn dataset is nearly saturated. Hence, a marginal improvement is obtained in the case of YUPenn dataset. It achieves perfect accuracy in nine out of fourteen classes. Overall, the accuracy of the proposed approach exceeds that of the state of the art methods, BoSE \& C3D by 2.14\% \& 1.66\% respectively for YUPenn dataset. 

\section{Conclusion}

As compared to the previous spatio-temporal approaches, we focus on capturing temporal variations of very powerful spatial descriptors provided by CNN. This method is computationally efficient than the traditional local feature extraction, encoding and pooling approaches. We observe that CNN spatial descriptors are excellent representatives of spatial information, as demonstrated by the accuracies obtained using only a single frame per video. This is due to the fact that most of the natural scenes, in spite of the inherent dynamism, are highly correlated with their spatial attributes. However, there is a large uncertainty in the performance, as it is highly dependent on which frame is chosen from the video. We propose methods that utilize multiple frames to improve accuracy as well as reduce this uncertainty to a large extent, as shown in Figure \ref{frame_vs_accuracy}. We also show that CNN pre-trained on hybrid of ImageNet and Places datasets outperforms the models trained on either of them alone. Thus, hybrid model provides very powerful representations for dynamic scene classification tasks.

We evaluate our algorithm on two standard and publicly available datasets (Maryland and YUPenn) for three pre-trained CNN models. Our proposed algorithm shows outstanding performance over the current state-of-the-art methods by 16.16 \% for Maryland and 1.66 \% for YUPenn datasets. The approach works well even for the very challenging Maryland dataset having large camera motion and jitter. High accuracies obtained for the various statistical moments indicate that similar classes have similar temporal statistics and that the spatial features temporally evolve in a similar way. Hence in future, various probabilistic methods can be explored by considering the joint distribution of the 4096 random variables (from the activation features of CNN) and finding out different models for different classes. We also report difficulty in applying vectorial pooling methods to datasets of such small sizes, as we find them not to give good results when compared to other aggregation approaches discussed in the paper. Overcoming this problem by dataset augmentation and vocabulary adaptation from another dataset is another future task. Expanding the approach to classify datasets with a large number of categories is a challenge which also needs to be investigated. 

\section*{Acknowlegdment}
The authors would like to thank SERB-DST, India for supporting this research through a start-up research grant.

%\begin{biography}{{\includegraphics[width=25mm,height=32mm,clip,keepaspectratio]{passport-visa photo.JPG}}}
%Name ... was born in ... more text.
%\end{biography} 

\section*{Authors}
\parpic{\includegraphics[width=1in,clip,keepaspectratio]{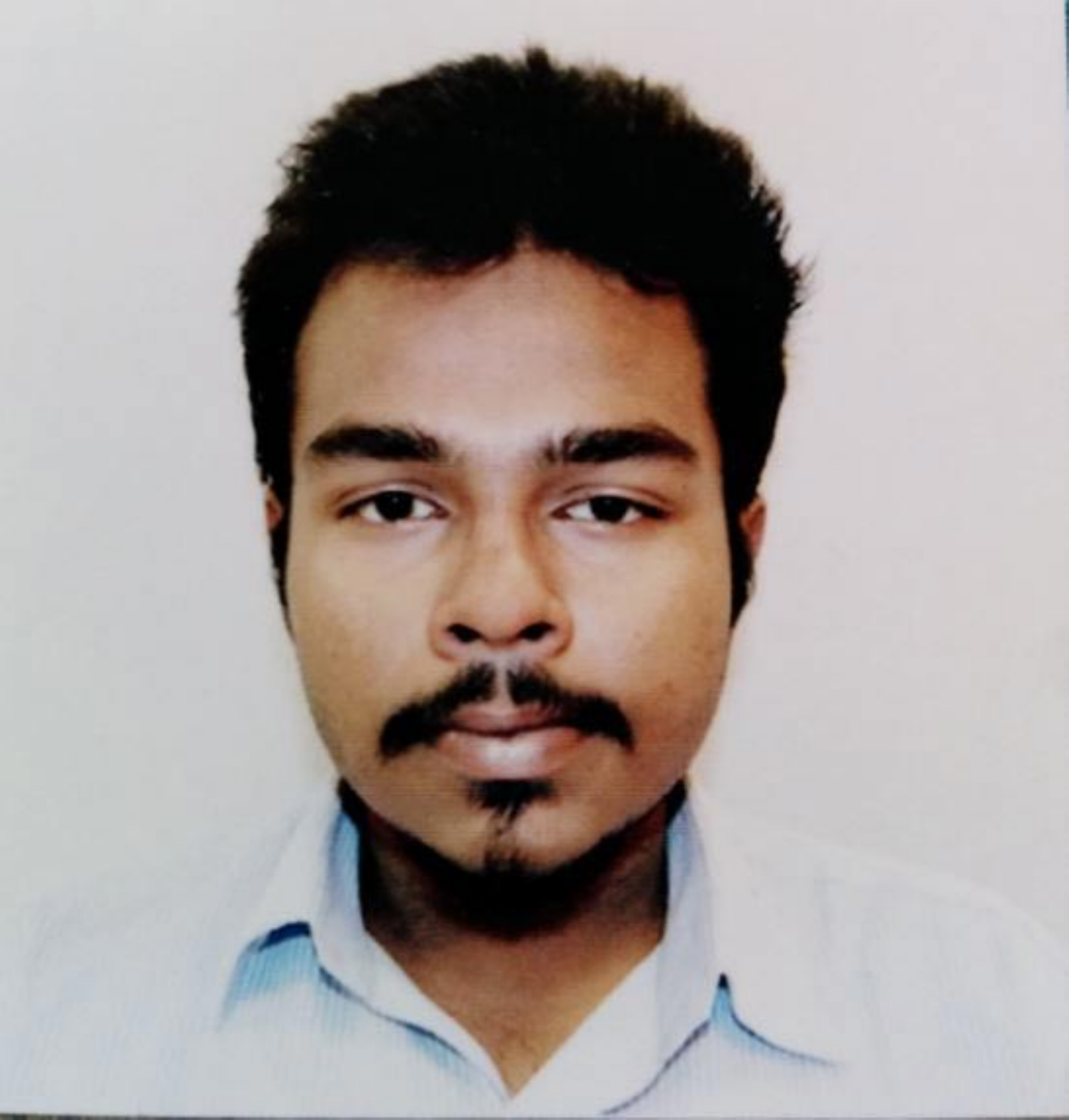}}
\noindent {\bf Aalok Gangopadhyay}\\ Aalok Gangopadhyay is an undergraduate student at Indian Institute of Technology Gandhinagar, majoring in Electrical Engineering. His research interests include Computer Vision and Machine Learning.\\

\parpic{\includegraphics[width=1in,clip,keepaspectratio]{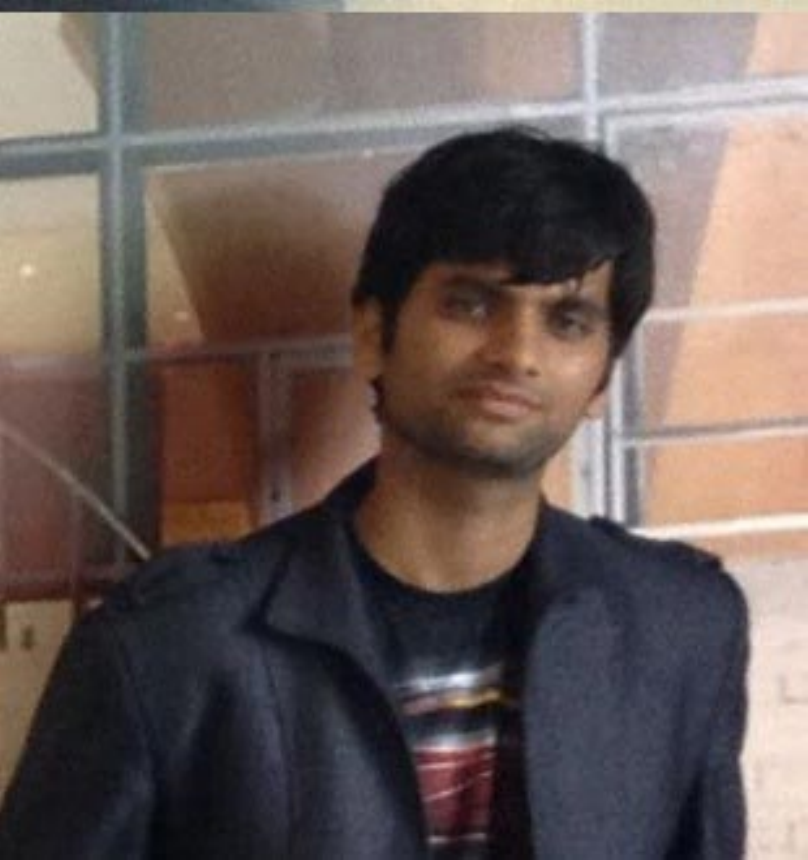}}
\noindent {\bf Shivam Mani Tripathi}\\ Shivam Mani Tripathi is an undergraduate student at Indian Institute of Technology Gandhinagar, majoring in Electrical Engineering. His research interests span across machine learning and computer vision.\\

\parpic{\includegraphics[width=1in,clip,keepaspectratio]{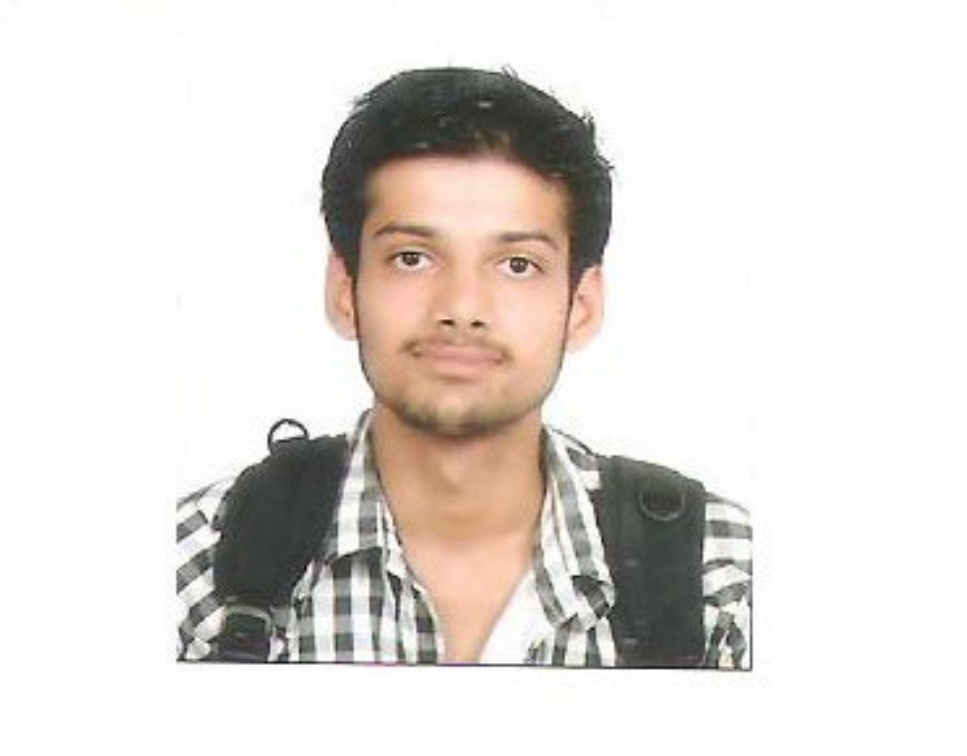}}
\noindent {\bf Ishan Jindal}\\
Ishan Jindal is currently a Junior Research Fellow at the Indian Institute of Technology Gandhinagar under Prof. Shanmuganathan Raman. He received his M.Tech. degree from Indian Institute of Technology Roorkee. His research interests are 3D reconstruction, Computer Vision,  Image and Video Processing and Machine Learning. He was also awarded DAAD Masters Scholarship for the year 2013-14.\\

\parpic{\includegraphics[width=1in,clip,keepaspectratio]{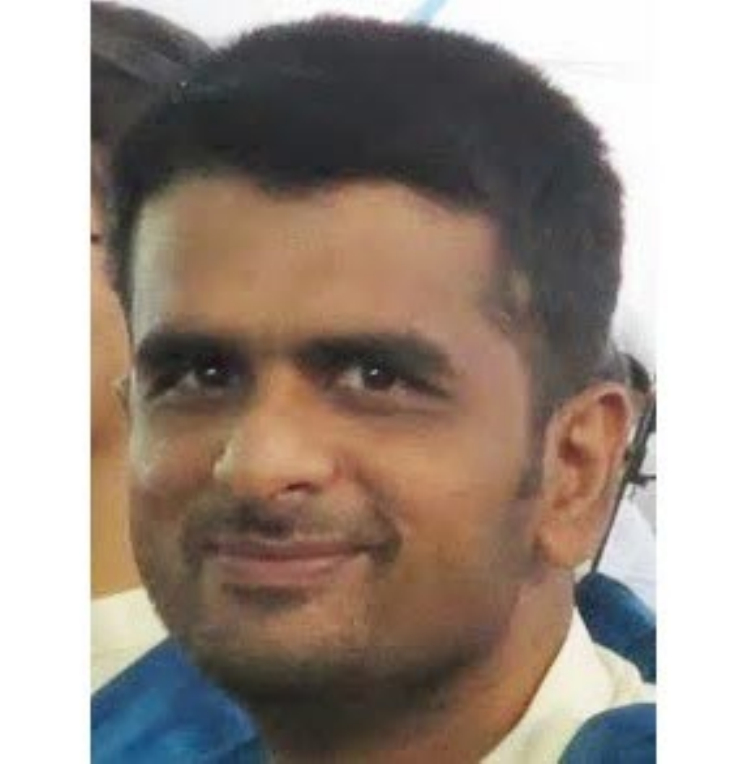}}
\noindent {\bf Shanmuganathan Raman}\\ Shanmuganathan Raman is currently an Assistant Professor of Electrical Engineering at the Indian Institute of Technology Gandhinagar. Prior to this position, he was an Assistant Professor at the Indian Institute of Technology Jodhpur. Prior to that he was a Postdoctoral Research Associate at the Indian Institute of Science, Bangalore. He received his B.E. and M. Tech. degrees from PSG College of Technology, Coimbatore and the Indian Institute of Technology Bombay, respectively. He obtained his Ph.D. degree from the Indian Institute of Technology Bombay. His research interests include Computer Vision, Computer Graphics, and Computational Photography. He was awarded Microsoft Research India Ph.D. Fellowship for the year 2007.


\section*{References}
\begin{thebibliography}{10}
\expandafter\ifx\csname url\endcsname\relax
  \def\url#1{\texttt{#1}}\fi
\expandafter\ifx\csname urlprefix\endcsname\relax\def\urlprefix{URL }\fi
\expandafter\ifx\csname href\endcsname\relax
  \def\href#1#2{#2} \def\path#1{#1}\fi


\bibitem{xiao2010sun}
J.~Xiao, J.~Hays, K.~A. Ehinger, A.~Oliva, A.~Torralba, Sun database:
  Large-scale scene recognition from abbey to zoo, in: Computer vision and
  pattern recognition (CVPR), 2010 IEEE conference on, IEEE, 2010, pp.
  3485--3492.

\bibitem{Krizhevsky_imagenetclassification}
A.~Krizhevsky, I.~Sutskever, G.~E. Hinton, Imagenet classification with deep
  convolutional neural networks, in: NIPS, p. 2012.

\bibitem{karpathy2014large}
A.~Karpathy, G.~Toderici, S.~Shetty, T.~Leung, R.~Sukthankar, L.~Fei-Fei,
  Large-scale video classification with convolutional neural networks, in: IEEE
  CVPR, 2014.

\bibitem{tran2014c3d}
D.~Tran, L.~Bourdev, R.~Fergus, L.~Torresani, M.~Paluri, C3d: Generic features
  for video analysis, arXiv preprint arXiv:1412.0767.

\bibitem{jia2014caffe}
Y.~Jia, E.~Shelhamer, J.~Donahue, S.~Karayev, J.~Long, R.~Girshick,
  S.~Guadarrama, T.~Darrell, Caffe: Convolutional architecture for fast feature
  embedding, arXiv preprint arXiv:1408.5093.

\bibitem{ILSVRCarxiv14}
O.~Russakovsky, J.~Deng, H.~Su, J.~Krause, S.~Satheesh, S.~Ma, Z.~Huang,
  A.~Karpathy, A.~Khosla, M.~Bernstein, A.~C. Berg, L.~Fei-Fei, {ImageNet Large
  Scale Visual Recognition Challenge} (2014).
\newblock \href {http://arxiv.org/abs/arXiv:1409.0575}
  {\path{arXiv:arXiv:1409.0575}}.

\bibitem{Csurka04visualcategorization}
G.~Csurka, C.~R. Dance, L.~Fan, J.~Willamowski, C.~Bray, Visual categorization
  with bags of keypoints, in: In Workshop on Statistical Learning in Computer
  Vision, ECCV, 2004, pp. 1--22.

\bibitem{Grauman05thepyramid}
K.~Grauman, T.~Darrell, The pyramid match kernel: Discriminative classification
  with sets of image features, in: IEEE ICCV, Vol.~2, 2005, pp. 1458--1465.

\bibitem{JDSP10}
H.~J\'egou, M.~Douze, C.~Schmid, P.~P\'erez,
  \href{"http://lear.inrialpes.fr/pubs/2010/JDSP10"}{Aggregating local
  descriptors into a compact image representation}, in: IEEE CVPR, 2010, pp.
  3304--3311.
\newline\urlprefix\url{"http://lear.inrialpes.fr/pubs/2010/JDSP10"}

\bibitem{conf/cvpr/PerronninD07}
F.~Perronnin, C.~Dance, Fisher kernels on visual vocabularies for image
  categorization, in: Computer Vision and Pattern Recognition, 2007. CVPR'07.
  IEEE Conference on, 2007, pp. 1--8.

\bibitem{Wang10locality-constrainedlinear}
J.~Wang, J.~Yang, K.~Yu, F.~Lv, T.~Huang, Y.~Gong, Locality-constrained linear
  coding for image classification, in: IEEE CVPR, 2010, pp. 3360--3367.

\bibitem{Schmid06beyondbags}
C.~Schmid, Beyond bags of features: Spatial pyramid matching for recognizing
  natural scene categories, in: IEEE CVPR, 2006.

\bibitem{DBLP:journals/corr/GongWGL14}
Y.~Gong, L.~Wang, R.~Guo, S.~Lazebnik, Multi-scale orderless pooling of deep
  convolutional activation features, arXiv preprint arXiv:1403.1840.

\bibitem{DBLP:journals/corr/SermanetEZMFL13}
P.~Sermanet, D.~Eigen, X.~Zhang, M.~Mathieu, R.~Fergus, Y.~LeCun,
  \href{http://arxiv.org/abs/1312.6229}{Overfeat: Integrated recognition,
  localization and detection using convolutional networks}, CoRR abs/1312.6229.
\newline\urlprefix\url{http://arxiv.org/abs/1312.6229}

\bibitem{DBLP:journals/corr/ZeilerF13}
M.~D. Zeiler, R.~Fergus, Visualizing and understanding convolutional networks,
  in: Computer Vision--ECCV 2014, Springer, 2014, pp. 818--833.

\bibitem{DBLP:journals/corr/DonahueJVHZTD13}
J.~Donahue, Y.~Jia, O.~Vinyals, J.~Hoffman, N.~Zhang, E.~Tzeng, T.~Darrell,
  Decaf: A deep convolutional activation feature for generic visual
  recognition, arXiv preprint arXiv:1310.1531.

\bibitem{razavian2014cnn}
A.~S. Razavian, H.~Azizpour, J.~Sullivan, S.~Carlsson, Cnn features
  off-the-shelf: an astounding baseline for recognition, arXiv preprint
  arXiv:1403.6382.

\bibitem{MLS09}
M.~Marszalek, I.~Laptev, C.~Schmid, Actions in context, in: IEEE CVPR, 2009,
  pp. 2929--2936.

\bibitem{5539864}
N.~Shroff, P.~Turaga, R.~Chellappa, Moving vistas: Exploiting motion for
  describing scenes, in: IEEE CVPR, 2010.

\bibitem{vasudevan2013dynamic}
A.~B. Vasudevan, S.~Muralidharan, S.~P. Chintapalli, S.~Raman, Dynamic scene
  classification using spatial and temporal cues, in: IEEE ICCVW, 2013, pp.
  803--810.

\bibitem{conf/cvpr/DerpanisLDW12}
K.~G. Derpanis, M.~Lecce, K.~Daniilidis, R.~P. Wildes, Dynamic scene
  understanding: The role of orientation features in space and time in scene
  classification, in: IEEE CVPR, 2012, pp. 1306--1313.

\bibitem{DBLP:conf/cvpr/FeichtenhoferPW14}
C.~Feichtenhofer, A.~Pinz, R.~P. Wildes, Bags of spacetime energies for dynamic
  scene recognition, in: IEEE, 2014.

\bibitem{candes2011robust}
E.~J. Cand{\`e}s, X.~Li, Y.~Ma, J.~Wright, Robust principal component
  analysis?, Journal of the ACM (JACM) 58~(3) (2011) 11.

\bibitem{guyon2012robust}
C.~Guyon, T.~Bouwmans, E.-h. Zahzah, et~al., Robust principal component
  analysis for background subtraction: Systematic evaluation and comparative
  analysis, Principal Component Analysis, P. Sanguansat, Ed.

\bibitem{lin2010augmented}
Z.~Lin, M.~Chen, Y.~Ma, The augmented lagrange multiplier method for exact
  recovery of corrupted low-rank matrices, arXiv preprint arXiv:1009.5055.

\bibitem{CC01a}
C.-C. Chang, C.-J. Lin, {LIBSVM}: A library for support vector machines, ACM
  Transactions on Intelligent Systems and Technology 2 (2011) 27:1--27:27,
  software available at \url{http://www.csie.ntu.edu.tw/~cjlin/libsvm}.

\bibitem{theriault2013dynamic}
C.~Theriault, N.~Thome, M.~Cord, Dynamic scene classification: Learning motion
  descriptors with slow features analysis, in: IEEE CVPR, 2013, pp. 2603--2610.

\bibitem{feichtenhofer2013spacetime}
C.~Feichtenhofer, A.~Pinz, R.~P. Wildes, Spacetime forests with complementary
  features for dynamic scene recognition, BMVC, 2013.
  
\bibitem{zhou2014places}
Zhou, Bolei, et al. "Learning deep features for scene recognition using places database." Advances in Neural Information Processing Systems. 2014.

\bibitem{XuTrecvid2015}
Xu, Zhongwen, Yi Yang, and Alexander G. Hauptmann. A Discriminative CNN Video Representation for Event Detection. arXiv preprint arXiv:1411.4006 (2014).

\bibitem{jegouVlad2010}
H. J´egou, M. Douze, C. Schmid, and P. P´erez. Aggregating local
descriptors into a compact image representation. In Proc. CVPR,
2010.

\bibitem{vlfeat}
Vedaldi, Andrea, and Brian Fulkerson. VLFeat: An open and portable library of computer vision algorithms. Proceedings of the international conference on Multimedia. ACM, 2010.

\bibitem{sklearn}
Pedregosa, Fabian, et al. Scikit-learn: Machine learning in Python. The Journal of Machine Learning Research 12 (2011): 2825-2830.

\end{thebibliography}
\end{document}